\documentclass{article}
\usepackage{arxiv}
\usepackage[T1]{fontenc}
\usepackage[utf8]{inputenc}
\usepackage{units}
\usepackage{textcomp}
\usepackage{amsmath}
\usepackage{graphicx}
\usepackage{wasysym}
\usepackage[unicode=true]
 {hyperref}

\makeatletter
\usepackage{makeidx}

\makeatother

\begin{document}

\docdate{2024-04-11}
\institute{University of Stuttgart, 70569 Stuttgart, Germany, \email{eric.price@ifr.uni-stuttgart.de}\and Max Planck Institute for Intelligent Systems, 72076 Tübingen, Germany.}

\author{Eric Price\inst{1,2} \and Aamir Ahmad\inst{1,2} \thanks{We would like to acknowledge the work of Pranav Khandelwal in assisting the in-situ experiments as well as everyone at the Hortobágy National Park in Hungary who made this work possible. Our thanks also goes to Windreiter for a great airship, designed and built on very short notice.}}

\title{Airship Formations for Animal Motion Capture and Behavior Analysis}\maketitle

\begin{abstract}
Using UAVs for wildlife observation and motion capture offers manifold
advantages for studying animals in the wild, especially grazing herds in open
terrain. The aerial perspective allows observation at a scale and depth that is
not possible on the ground, offering new insights into group behavior. However,
the very nature of wildlife field-studies puts traditional fixed wing and
multi-copter systems to their limits: limited flight time, noise and safety
aspects affect their efficacy, where lighter than air systems can remain on
station for many hours. Nevertheless, airships are challenging from a ground
handling perspective as well as from a control point of view, being voluminous
and highly affected by wind.
In this work, we showcase a system designed to use airship formations to track,
follow, and visually record wild horses from multiple angles, including airship
design, simulation, control, on board computer vision, autonomous operation and
practical aspects of field experiments.
\keywords{Airship Formations, Cooperative Tracking, Active Perception, Wildlife Monitoring}
\end{abstract}

\section{Introduction}
\begin{figure}
\begin{centering}
\includegraphics[width=0.45\columnwidth]{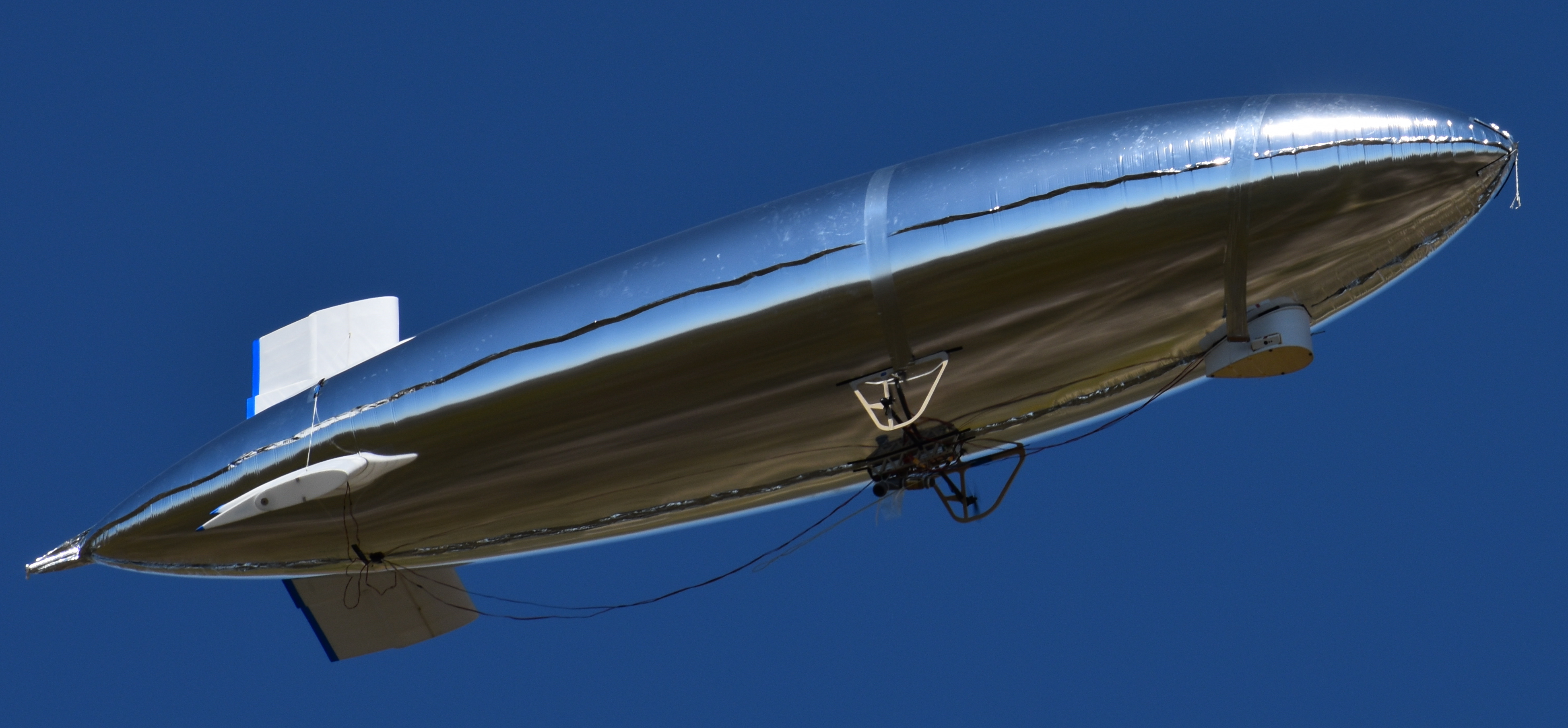}
\includegraphics[width=0.45\columnwidth]{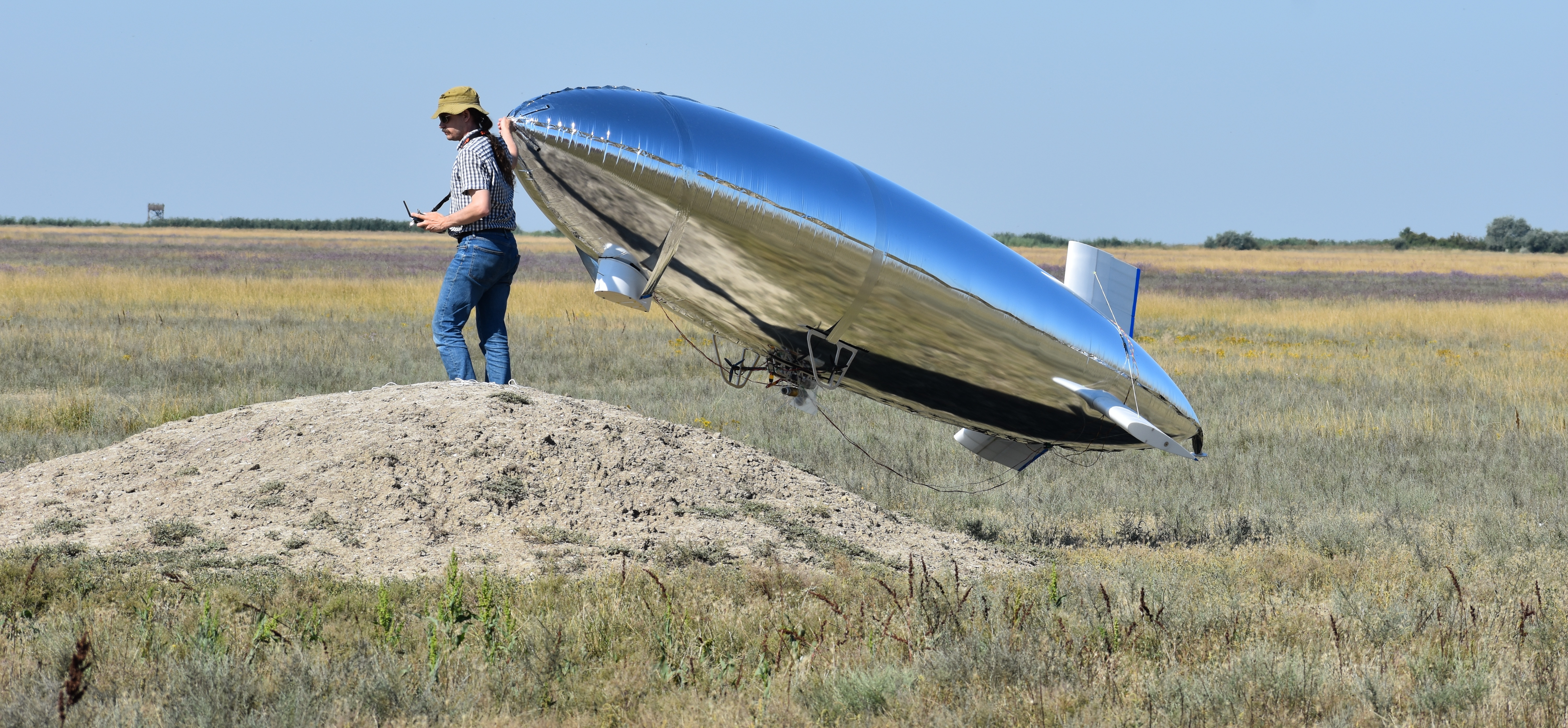}
\includegraphics[width=0.45\columnwidth]{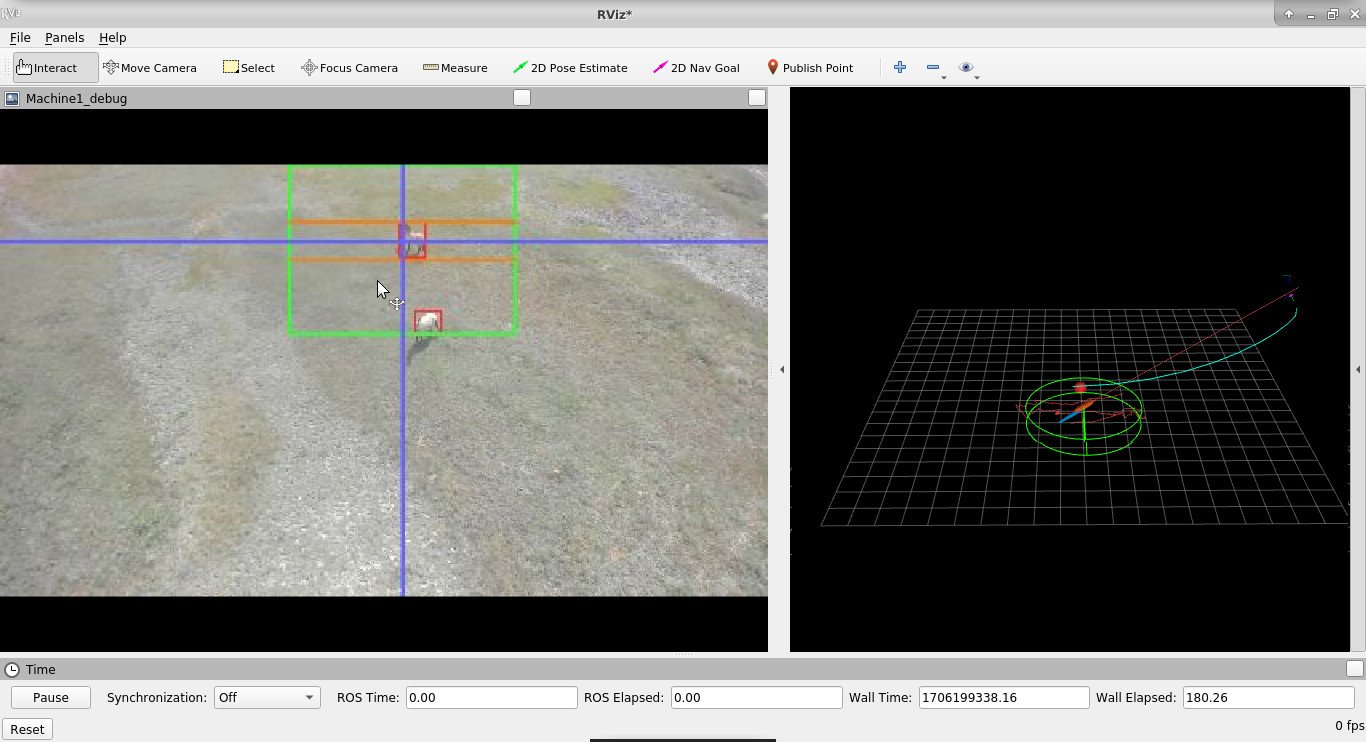}
\includegraphics[width=0.45\columnwidth]{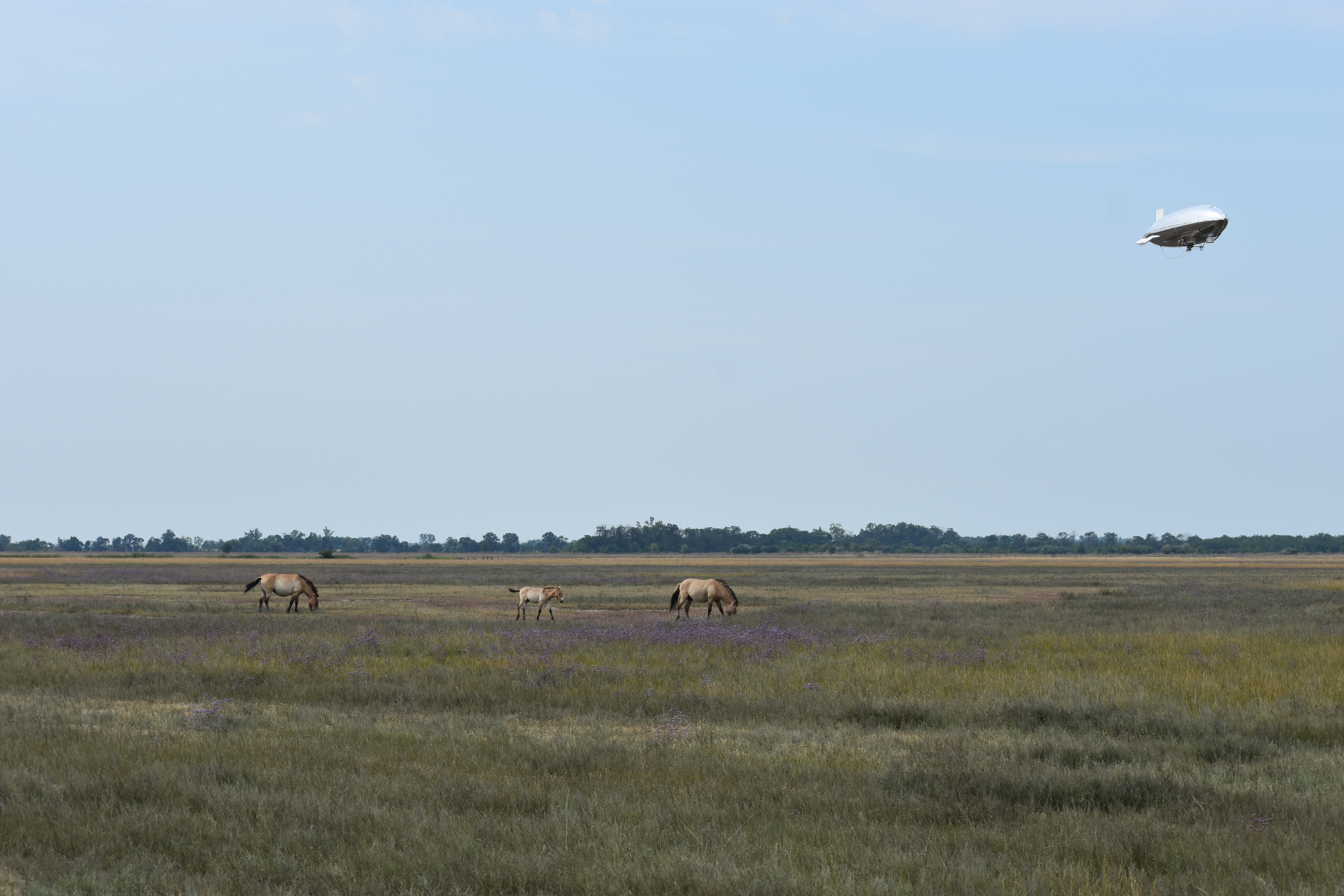}
\par\end{centering}
\caption{\label{fig:blimp_flying}Top left: Blimp prototype. Top right: Takeoff location.\newline
Bottom left: GUI for horse tracking. Bottom right: Airship observing horses.}
\end{figure}

In this work, we evaluate the use of lighter than air UAVs for use in studying
animals in the wild, especially grazing herds in open terrain. The aerial
perspective allows observation at a scale and depth that is not possible on the
ground and offers new insights into group behavior.  Airships seem like an
obvious choice for wildlife field-studies. Unlike fixed wing and multi-copter
systems, which suffer from limited flight time and have noise and safety
concerns, airships are characterised by static buoyancy and low mass density.
This makes them both inherently efficient and safe.  However, airships can be
challenging from a ground handling and control point of view. Airships can also
be more affected by wind than ``traditional'' UAVs.  In this work, we showcase
our system of lighter than air vehicles designed to track, follow and visually
record wild horses from multiple angles. We also show the results of
our field evaluations. The following aspects are of particular
importance for this airship application:

\subsection{Airship Design}
The design of an airship for animal studies is determined by the following
requirements: i) Speed and efficiency: the airship must be capable of
station-holding for extended times in open terrain, which frequently encounters
significant winds. On many days during our experiments, the wind was in excess
of $5 \mathrm{\frac{m}{s}}$ with gusts exceeding $10 \mathrm{\frac{m}{s}}$. An
airship build for extended flight times must be designed for comparable
sustained airspeeds, which requires a high fineness ratio, low drag and
efficient thrusters. ii) Maneuverability: being able to maintain stable camera
tracks of moving subjects on the ground, especially in the presence of wind and
turbulence requires sufficient control authority in combination with passive
stability. iii) Payload: in addition to actuation, on board computer vision and
image processing are required to perform autonomous observation of moving
subjects. This means, capable embedded compute hardware with non-negligible
weight and power consumption must be carried and supplied with power for an
extended time. iv) Durability: operation ``in the wild'' for an extended time-frame
requires that hull, structure and payload are sufficiently rugged to maintain
function and lift-gas retention in the presence of weather, turbulence and
handling related strain. iv) Safety: operating a UAV in a national park and/or
near wildlife requires high reliability of flight control and navigation, as
well as vehicle electronics and mechanics with inbuilt fail-safes. v) Pressure:
changes in air pressure and temperature over extended mission times cannot be
ruled out. Hull pressure must be maintained either with sufficient margins or
with active control (e.g. ballonets).

\subsection{Sensing and Computer Vision}
When studying wild animals and their behavior, a critical criteria is to be
non-invasive. We can not rely on markers, transmitters or similar strategies
requiring hardware on the animal.  Instead, passive sensors must be used. Since
the visual modality is most important for studying the subjects in question,
using vision for real-time sensing and tracking is the obvious choice.

\subsection{Autonomous Control}
To navigate a formation of airships near animals, both formation and individual
vehicle control needs to be addressed under consideration of both airship
dynamics, wind and motion of the subject animals. To address the needs of
directional sensors and computer vision, the control algorithm needs to keep
the subject animals in the camera field of view. This should be done from
multiple angles with multiple vehicles simultaneously, while also maintaining a
safe distance to avoid collisions between all vehicles, terrain or disturbances
to subjects.

\subsection{Simulation}
Complex autonomous operations and their algorithms need to be extensively
tested under realistically simulated conditions before fielding them. This
should include realistic (flight) physics, sensor behavior including sensor
noise, actuator effects and disturbances from the environment including wind
and turbulence. To minimize discrepancies between simulation and the real
world, software and hardware in the loop simulation is desired, where sensing
and control methods are agnostic w.r.t. the source of data, i.e. simulated vs.
real sensors. Timing/sequencing should be as realistic as possible.

\subsection{Field Experiments and Practical Considerations} Operating an
airship under field conditions can be more complicated than operating small
multi-copters, which can often be transported by a single person and, if
necessary, hand launched. The main limitations are the need to inflate the
airship with lift-gas and the significant volume and drag of the vehicle
once inflated. Inflating and trimming the vehicle therefore should ideally
happen under sheltered conditions. While making the vehicle flight-ready, it is
very vulnerable to wind and turbulence, especially near obstacles. Having a way
to arrest both the nose and the tail of the vehicle securely during handling is
important.  Takeoff and landing are another challenge, especially in vehicles
that do not have vertical thrust.  Thermal activity and detachment of hot air
bubbles are common phenomena in open terrain, especially on sunny days.
Typically, airship UAVs are trimmed heavier than air by up to 10\% of their mass
to ensure the vehicle safely returns to the ground in case of control-loss.
This can be insufficient when the airship encounters thermal updrafts. Even
high thrust in combination with a nose-down attitude might not be sufficient to
maintain altitude. Similarly, micro bursts with significant down-force can be
encountered, which can threaten to drive an airship into the ground. A
combination of navigation and control strategies, as well as human override
capability can be utilized to overcome these challenges.  It is crucial that
all operators of airship UAVs are familiar with these phenomena and how to
handle them.

\section{State of the art\label{sec:stateoftheart}}

While the benefits of UAVs for wildlife research are manifold and undisputed
\cite{Mazumdar2022,doi:10.1139/juvs-2015-0021}, multi-copters have known
drawbacks including disturbance of animals
\cite{Bennitt2019,10.7882/AZ.2021.015} and limited flight time \cite{9691840}.

A recent literature study reported a significant increase in use of lighter
than air vehicles for wildlife research \cite{doi:10.1139/dsa-2023-0052} citing
several advantages, including significantly higher efficiency, lower cost and longer
flight times. However, established  systems used in the field today are tethered
aerostats without propulsion or autonomous mobility
\cite{cf011db3-3abf-38f1-84f4-d45844a77116,10.7882/AZ.2020.004}. 

Manned Airships have been extensively used for aerial observation up to the
1930's, including geographic aerial surveys and military observation in World
War I \cite{cross1993zeppelins}. The use of airships for research expeditions in
Africa had been proposed as early as 1908 \cite{Frobenius1913}. Airships
have been used for multiple research expeditions, mostly into the arctic
\cite{nelson1993airships}. However, modern manned designs are too expensive to
operate to compete with unmanned solutions, especially in remote areas.

Using autonomous robotic airships for environmental monitoring and wildlife
observation has been proposed as a general concept
\cite{doi:10.1139/juvs-2015-0021}, as well as motivation for both ship design
proposals \cite{10.1007/978-3-642-33161-9_69} and airship component research
\cite{6959391}. However, to our knowledge, no prior work described the design of
a robotic airship specifically for this purpose nor its evaluation for wildlife
observation under realistic field conditions.

This work builds on an extensive foundation of airship simulation and modelling
\cite{10.1007/978-3-030-95892-3_46}, computer vision and state estimation
\cite{eprice2018dnncoopvistrack} and model predictive control for airships in
formations relative to observation subjects \cite{10092932}.

\section{Methodology}

\subsection{\label{airshipdesign}Airship Design}
 
\begin{figure}
\begin{centering}
\includegraphics[width=0.99\columnwidth]{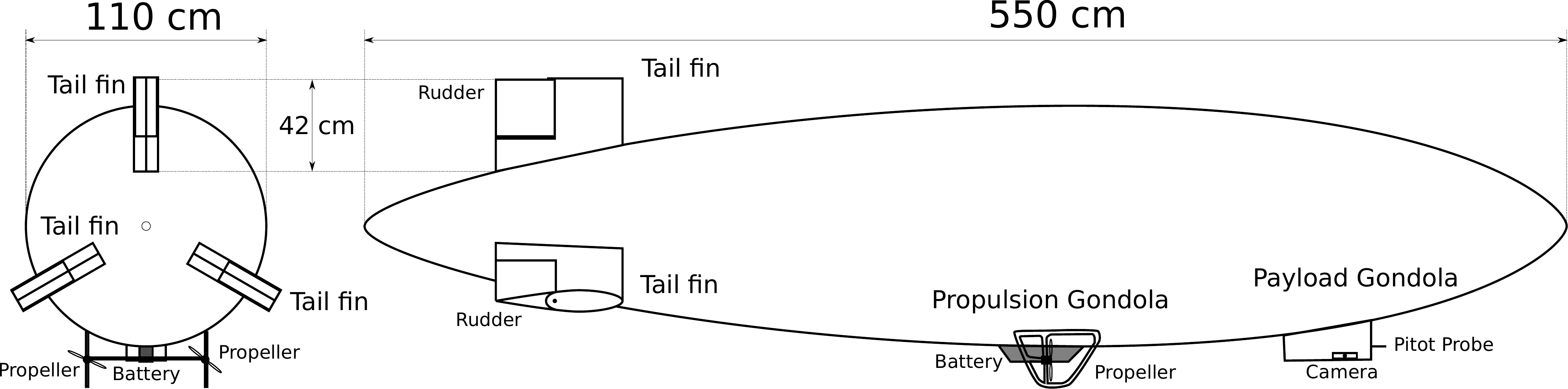}
\end{centering}
\caption{\label{fig:blimp_measures}The dimensions and physical layout of the airship prototype, including rudders, propulsion and payload gondola.}
\end{figure}

\subsubsection{Structure}
In cooperation with a commercial partner, German company ``Windreiter''
\cite{windreiter}, we designed a $5.5\mathrm{m}$, $3.6 \mathrm{m^3}$ single
hull prototype. It has a metal coated polymer film envelope with a mass of
$430\mathrm{g}$. To minimize drag, a configuration with 3 tail-fins was chosen.
The propulsion gondola with batteries is located under the center of mass.
Additionally, the airship is equipped with a forward mounted payload gondola,
to physically distance sensors such as the magnetometer and pressure sensors
from propulsion disturbances.  The fins, rudders and gondolas are made from
depron and plywood-depron sandwich. Each fin has a mass of approx
$200\mathrm{g}$. The total maximum payload capacity at sea level is
$1800\mathrm{g}$ (Fig. \ref{fig:blimp_flying} top left, Fig.
\ref{fig:blimp_measures}). 
A rope for ground handling has been attached to the nose of the vehicle.

\subsubsection{Power Supply and Propulsion}
The airship has been equipped with 2 $4\mathrm{S}$ $5500 \mathrm{mAh}$ Lithium
Polymer batteries ($125 \mathrm{Wh}$), with the option of flying with a third
battery for $186 \mathrm{Wh}$. Thrust is generated by 2 fixed, brush-less
propellers in pull configuration, attached to the propulsion gondola. Power and
propulsion have been designed w.r.t the air-frame with the goal of a $12
\mathrm{\frac{m}{s}}$ top speed and sustained flight at $8
\mathrm{\frac{m}{s}}$ for at least one hour.

\subsubsection{Payload and Avionics}
We equipped the airship with 2 Logitech Brio 4k USB webcams
\cite{logitechbrio}, one of which is front-facing for navigation while the
other is pointing to starboard and 30 degrees down for subject observation. The
main computer is an NVIDIA Jetson TX2 \cite{nvtx2datasheet} on a AUVIDEA J-120
embedded board \cite{j120datasheet}. We utilize the TX2's inbuilt 5 Ghz Wifi
for inter-vehicle communication and communication with a ruggedized
ground-station laptop. A Graupner HoTT transceiver \cite{graupnerhottdatasheet}
is used for remote flight control and long range telemetry operating on 2.4
GHz. The flight controller (FC) is an Openpilot Revolution STM32 based embedded
board \cite{archivedoprevo}, equipped with IMU, 3 axis magnetometer and
barometric static pressure sensor. Additionally, we equip a GPS receiver and a
dynamic pressure based airspeed sensor consisting of a pitot tube and a
differential pressure sensor. The TX2 is equipped with a $1 \mathrm{TB}$ SSD
drive for on board data storage, sufficient for several hours of recordings.

\subsection{\label{method:sense:CV}Sensing and Computer Vision}
The flight controller (FC) runs our ROS-enabled \cite{quigley2009ros} version
of Librepilot \cite{librepilotrepo}. It includes an Extended Kalman Filter
(EKF) for state estimation which fuses GPS, IMU, barometer and magnetometer for
self-pose estimate (position and orientation), which is made available to ROS
running on the TX2 main computer.  We use our own distributed cooperative
Bayesian foveated visual tracker \cite{eprice2018dnncoopvistrack} across one or
multiple vehicles to improve the self pose estimate and track a motion capture
(MoCap) subject, i.e. an animal on the ground. For visual detection in the
relevant field of view (fov), our approach uses a convolutional single shot
multibox detector (SSD) \cite{10.1007/978-3-319-46448-0_2} running on the TX2.

\subsection{Autonomous Control}
All flight controls, i.e. all 3 rudders and the 2 main thrusters are directly
controlled by the FC. A cascaded PID controller is implemented on the FC, which
allows basic low level control, i.e. flying at a controlled airspeed and
climb/sink rate on a straight or curved trajectory with controlled rotational
rate.  Furthermore, it is capable of position hold and waypoint navigation from
our previous work \cite{10.1007/978-3-030-95892-3_46}. This low-level
controller is also used to enforce a ``sky-box'', a geo-fenced area in which more
sophisticated autonomous vision based navigation is permitted. This ensures no
ground collisions or fly-away conditions are encountered, even if there is a
failure in the high level control. Within the sky-box, when in autonomous mode,
a Model Predictive Controller (MPC) \cite{10092932} estimates the wind vector
based on the airship's groundspeed, dynamic pressure (i.e. airspeed), rotation
rate and estimated angle of attack. The MPC knows the camera orientation on the
airship.  It calculates the optimal trajectory to keep the subject, which is
tracked as described in Subsec. \ref{method:sense:CV}, centered in the camera
field of view, while maintaining equal angular separation w.r.t. the subject
between all vehicles in the formation. For this, optimal trajectories for all
vehicles are calculated.  The MPC takes collision avoidance constraints into
account for both moving vehicles, the subject and known static obstacles. It
avoids direct overflight of the MoCap subject for safety reasons.
In the absence of obstacles and if winds are steady, the resulting trajectories
resemble circular orbits around the subject, which are offset to one side based
on the subjects velocity and/or wind \cite{10092932}. In strong winds,
different solutions are possible. A single airship can ``hover'' with its nose in
the wind and/or fly parallel to a fast moving subject.
Manual override via the remote control transmitter is always possible for
safety reasons.

\subsection{Simulation}

We implemented an aerodynamic physics simulation of airship formations based
on our previous work of simulation of non-rigid airships in turbulent wind
\cite{10.1007/978-3-030-95892-3_46} using ROS and Gazebo \cite{Gazebo}. This
includes hardware in the loop simulation, in which the low level flight control
algorithms are running on a physical or software emulated FC in real-time.
Since our formations are subject-centered, we include a simulated subject,
which the SSD detector can detect in simulated camera images as presented in
our simulation of MoCap using a formation of drones
\cite{eprice2018dnncoopvistrack} and also for our MPC evaluation
\cite{10092932}. This allows for real-time simulation of all components under
diverse conditions, including different wind and turbulence regimes, as well as
the loop-closure between perception, sensor fusion and formation control
relative to the tracked subject. All our control and simulation code is
available open source for ease of reproduction and the benefit of the community
\cite{coderepo}.

\section{Field Experiments, Practical Considerations and Results}

In August 2023, we conducted field experiments at the Hortobágy National Park
in Hungary, which is an UNESCO world heritage property \cite{hortobagy}. Our
intent was to develop procedures and to assess the efficacy of airship and
airship formations for motion capture and behavioral study of wild horses,
particularly Przewalski's horses (Equus ferus przewalskii). We used manually
annotated aerial drone footage from the same herd of Przewalski's taken the
previous year to train our SSD detector network, in order to reliably detect
these horses in aerial video frames. We brought the previously un-flown
airship prototype (Subsec. \ref{airshipdesign}) to the site in an off-road
vehicle, along with two $50 \mathrm{l}$ $200 \mathrm{Bar}$ helium lift-gas
cylinders (sufficient for $20 \mathrm{m}^3$).  The airship was assembled
on-site. Although the use of a gazebo (tent) was considered, we opted instead
for an empty farm-shed which was available on site to assemble and fill the
airship. To protect the sensitive hull from dirt, a tarp was placed on the
ground. A net over the hull was used to prevent the hull from rising until
attachments, batteries and sufficient ballast could be equipped. A pressure
reducer valve and a hose were used to fill the airship with helium. Metal
screws in a plastic bag were attached to the main gondola as ballast, to trim
the airship approx.  $300 \mathrm{g}$ heavier than air.  Bringing the airship
to the takeoff location was challenging due to gusty winds. It was not possible
to maintain handling control over the airship using the gondola as a holding
point due to excessive lateral wind forces. Handling became manageable using a
rope attached to the nose. Takeoff and landing were performed by releasing and
catching the airship by hand on the nose and front gondola. This was done on a
small hill to increase ground clearance (Fig.  \ref{fig:blimp_flying} top
right).

\begin{figure}
\begin{centering}
\includegraphics[width=0.99\columnwidth]{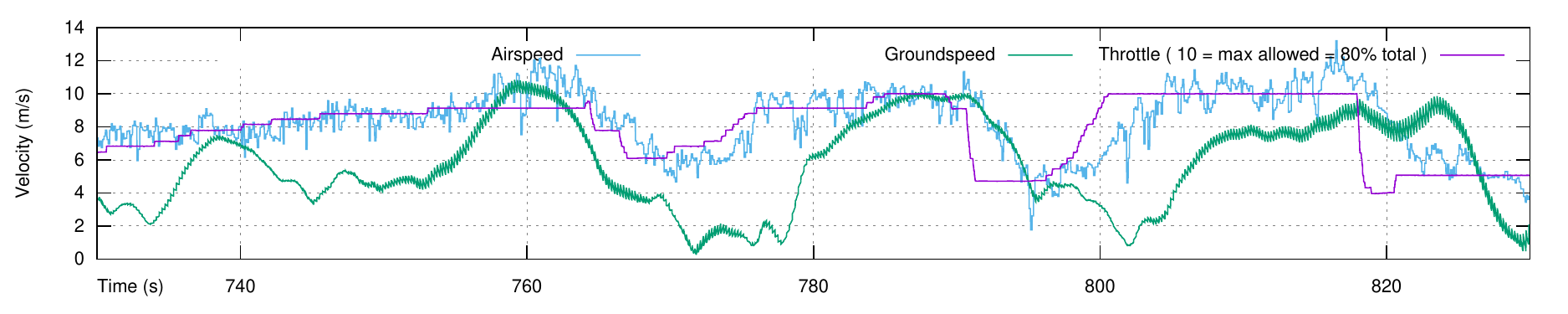}
\end{centering}
\caption{\label{fig:blimp_speed}Logged airspeed vs groundspeed in
back-and-forth flights on the same track (ROS). Peak power consumption was
$333\mathrm{W}$ at maximum throttle (not displayed).}
\end{figure}

The vehicle maiden flight and initial tuning of the FC control coefficients
were all conducted in situ. For safety reasons, the maximum throttle of the
propellers was limited to $80\%$, at which an airspeed of $11
\mathrm{\frac{m}{s}}$ was achieved (Fig. \ref{fig:blimp_speed}). We measured a
peak current of $23\mathrm{A}$ at sustained $80\%$ throttle in flight with a
battery voltage of $14.5\mathrm{V}$ resulting in a maximum sustained power
consumption of $333\mathrm{W}$. Power consumption at $40\%$ throttle (half the
allowed) was $6.9\mathrm{A}$ or $100\mathrm{W}$ at approx
$6\mathrm{\frac{m}{s}}$. We did not measure engine RPM. The expected flight
endurance is reciprocal with electric current and can be calculated based on our
$10\mathrm{Ah}$ battery capacity. Current and power are approximately
proportional to the square of both the throttle setting and measured airspeed.
Over several test-flights the various components were evaluated, including the
formation controller using a 1-airship formation around a simulated stationary
subject.  It was determined that a minimum airspeed of $4 \mathrm{\frac{m}{s}}$
was necessary to maintain sufficient vertical control authority in the presence
of both strong thermals and micro bursts. Once the control accuracy was
sufficient, the vehicle was flown manually into the vicinity of wild horses. The
subject tracking and formation control was then activated in flight for an
autonomous formation consisting of 1 airship and one subject horse. The initial
acquisition of the subject in the camera image was challenging, since the
camera was mounted sideways and has a limited field of view. It was difficult
for the pilot to estimate the relative position of the airship in the air and
the horses on the ground, especially w.r.t. relative distance. We later added a
forward facing navigation camera to the gondola to address this problem. Both
cameras can be monitored from the ground in real time.  The formation was
monitored with a ruggedized ground-station laptop, connected to the airship via
Wifi. We achieved a practical Wifi range of approximately $150\mathrm{m}$,
which was a limiting factor but sufficient for our experiment. The operator
could see a low-res visualization of the tracked subject, as seen by the
on-board camera, on the laptop screen. We used the ``rviz'' ROS tool with an
interactive camera plugin to point and click on the correct horse to track,
which re-initialized the tracker with the projected camera coordinate. This was
sufficient to initialize the formation control once the animal was in sight. A
3D-visualization showed the airships relative to the subject, along with the
estimated position uncertainty (Fig. \ref{fig:blimp_flying} bottom left).  We
collected video data of the horses with the autonomous airship in very
challenging conditions with approx $6-8 \mathrm{\frac{m}{s}}$ winds (Fig.
\ref{fig:blimp_flying} bottom right). Video and telemetry data of the flight
were stored on the on-board SSD, both in form of compressed raw-video,
radio-telemetry log files and ROS-bag-files. The used video storage format was
mjpeg compressed raw video data as it came from the camera, enriched with
microsecond accurate frame timing information. Telemetry was stored as a
telemetry stream-dump with timing information, allowing replay and analysis in
the Librepilot GCS software.  Recorded ROS data consisted of crucial
airship-state information including position, velocity, IMU data, airspeed as
well as remote control inputs and setpoints at a rate of $500\mathrm{Hz}$. ROS
data also included low resolution video frames with annotations for debugging.

\subsection{Discussion}

The airship prototype exceeded our expectations. We went from maiden flight to
successful autonomous flight tracking horses within only 4 days. The
experiments showed that a vehicle with sufficient maximum airspeed and
endurance has no problems tackling challenging wind conditions. Based on measured
airspeed and power consumption, the calculated endurance of our airship at
$8\mathrm{\frac{m}{s}}$ would be $50 \mathrm{min}$ which is technically below
the targeted $1 \mathrm{hour}$. However, we attribute a large part of the high
power consumption to drag caused by a foil-belt used for payload-gondola
attachment, which started twisting and fluttering at high airspeeds.  We flew
with approximately $600\mathrm{g}$ of ballast, which could be partially
replaced by another $5000\mathrm{mAh}$ battery resulting in a $50\%$ endurance
increase. Unlike multi-copters, adding more batteries to airships does not
result in higher power consumption. This indicates that minor modification to
the existing design would result in both higher flight speeds and an increased
endurance, meeting our design goals.
The ultra-light metal-coated polymer hull is fragile in ground handling and can
easily be scratched, punctured or otherwise damaged. We also found out that
touching the hull with sweaty hands causes corrosion and discoloration of the
metal coating. A single hull vehicle also does not allow for dynamic pressure
compensation without adding lift-gas. A double hull, although more limiting on
payload, would be better suited in the long term, and allow easy installation
of an air filled ballonet. Ground handling was also the limiting factor w.r.t.
encountered wind speeds and gusts, since bringing the airship into a hangar,
barn or shelter without colliding is very hard when there is lateral wind.
During our experiments, ground handling mishaps were the only causes of damage
to the vehicle, especially cracks in the depron material, which could be
repaired in situ. Small punctures in the hull, caused by thorny vegetation,
were easily fixed with patches of sticky tape, without noticeable loss of lift
gas.  Thermals and downdrafts were initially concerning, but could easily be
overcome by ensuring the airship always maintained sufficient airspeed, i.e.
vertical control authority.  The system as a whole showed that airships are a
feasible and convenient tool for autonomous animal observation and research,
matching and - in the aspect of endurance - easily exceeding the abilities of
drones. The perceived propulsion noise of the airship was also significantly
lower than noise of a drone of comparable payload capacity. According to
assessments by wildlife researchers observing our experiments, the presence of
the airships did not disturb the horses.

\section{Conclusions}

We developed a system that allows for a formation of airships to autonomously
follow and record horses from an aerial view, enabling unmatched research
opportunities such as motion capture and behavior analysis. We showed that
using airships for such studies in situ and in the wild is feasible. At a minimum a
large tent or similar shelter is needed to fill and fully assemble airships
and/or keep them out of the elements between experiments. This overhead is not
unreasonable compared to other aerial vehicles, and we determined the benefits
outweigh the inconveniences. Takeoff and handling in the air are not harder
than other drones, and the data collected is of equal quality or superior. The
main benefit of airships is the long flight time on the order of magnitude of hours, which
allow for extended studies, impossible with multi-rotor drones. Control of
airships, although more challenging than multi-copters capable of
omni-directional movement can be solved with more sophisticated models and
control methods, as we have shown.  Our work is based on both open and
proprietary hardware designs which are commercially available as well as
publicly available open source projects. To foster lighter than air research
and use of airships especially for wildlife research, we have shared all our
simulation and control code, developed in previous works
\cite{eprice2018dnncoopvistrack,10.1007/978-3-030-95892-3_46,10092932}, for the
benefit of the
community\footnote{\url{https://github.com/robot-perception-group/Airship-MPC}}.


 \bibliographystyle{splncs04}
\phantomsection\addcontentsline{toc}{section}{\refname}\bibliography{paper}

\end{document}